\documentclass[runningheads]{llncs}
\usepackage{graphicx}
\usepackage{hyperref}
\graphicspath{ {./images/} }
\begin{document}

{\let\thefootnote\relax\footnotetext{Copyright \textcopyright\ 2021 for this paper by its authors. Use permitted under Creative Commons License Attribution 4.0 International (CC BY 4.0).}}

\title{Building and Evaluating Universal Named-Entity Recognition English corpus}
%
%
\author{Diego Alves\orcidID{0000-0001-8311-2240}\and
Gaurish Thakkar\orcidID{0000-0002-8119-5078} \and
Marko Tadić\orcidID{0000-0001-6325-820X}}
\authorrunning{D. Alves et al.}
%
\institute{Faculty of Humanities and Social Sciences, University of Zagreb, Zagreb 10000, Croatia \\ \email{\{dfvalio,marko.tadic\}@ffzg.hr, gthakkar@m.ffzg.hr}}
\maketitle              
\begin{abstract}
This article presents the application of the Universal Named Entity framework to generate automatically annotated corpora. By using a workflow that extracts Wikipedia data and meta-data and DBpedia information, we generated an English dataset which is described and evaluated. Furthermore, we conducted a set of experiments to improve the annotations in terms of precision, recall, and F1-measure. The final dataset is available and the established workflow can be applied to any language with existing Wikipedia and DBpedia. As part of future research, we intend to continue improving the annotation process and extend it to other languages.

\end{abstract}
%
%
%
\section{Introduction}

Named entity recognition and classification (NERC) is an important field inside Natural Language Processing (NLP), being a crucial task of information extraction from texts. It was first defined in 1995 in the 6th  Message  Understanding  Conference  (MUC-6)\cite{chinchor-robinson-1998-appendix} and since then has been used in multiple NLP applications such as events and relations extraction, question answering systems and entity-oriented search. 

As shown by Alves et al.\cite{alves2020uner}, NERC corpora and tools present an immense variety in terms of annotation hierarchy and formats. NERC hierarchy structure is usually locally defined considering the final NLP application where it will be used at. If certain types like "Person", "Location", and "Organization" are present in almost every NERC system, some corpora are composed of more complex types of annotation. It is the case, for example, of Portuguese Second HAREM\cite{freitas2010second}, Czech Named Entity Corpus 2.0\cite{vsevvcikova2007named} and Romanian Ronec\cite{DBLP:journals/corr/abs-1909-01247}. Multilingual alternatives also exist, it is the case of spaCy software\cite{spacy} proposing two different single-level hierarchies composed of either 18 or 4 NERC types following OntoNotes 5.0\cite{AB2/MKJJ2R_2013} and Wikipedia\cite{journals/ai/NothmanRRMC13} respectively.

Unlike Part-of-Speech and Dependency Parsing tagging which have Universal Dependencies\footnote{https://universaldependencies.org}, there is no universal alternative for NERC in terms of annotation framework and multilingual repository following the same standards.

Hence, we use the Universal Named Entity (UNER) framework which is composed of a complex NERC hierarchy inspired by Sekine's work\cite{sekinewebsite} and propose a process which parses data from Wikipedia\footnote{https://www.wikipedia.org/}, extract named entities through hyperlinks, aligns them with DBpedia\footnote{https://www.dbpedia.org/}\cite{madoc37476} entity classes and translates them into UNER types and subtypes. This process can be applied to any language present in both Wikipedia and DBpedia, therefore generating multilingual NERC corpora following the same procedure and hierarchy. 

Thus, UNER is useful for multilingual NLP tasks which need recognition and classification of named entities going beyond classical NERC hierarchy involving only a few types. UNER data can be used in its totality or can be easily adapted to specific needs. For example, considering the classic “Location” type usually present in NERC corpora, UNER can be used to get more detailed information: if an entity is more specifically a country, mountain, island, etc.

This paper presents the UNER hierarchy and its workflow for data extraction and annotation. It details the application of the proposed process on English language with qualitative and quantitative evaluation of the automatically annotated data. It also presents the evaluation of different alternatives implemented for the improvement of the generated dataset. 

This article is organized as follows: In Section 2, we present the state-of-the-art concerning NERC automatic data generation workflows; in Section 3, we describe UNER framework and hierarchy, and in Section 4 the details of the data extraction and annotation workflow and evaluation of the generated dataset. In Section 5, we report the experiments that were conducted for improving the dataset quality in terms of Precision and Recall. Section 6 is dedicated to the discussion of the results, and in Section 7, we present our conclusions and possible future directions for research.

\section{Related Work}

UNER framework was first introduced by us in a previous article \cite{alves-etal-2020-evaluating}, where we defined its hierarchy. In this article, the framework was revised and a workflow for automatic text annotation was developed and applied to generate an annotated corpus in English with the respective evaluation.

Deep learning has been employed in NERC systems in recent years improving stat-of-the-art performance, and, therefore, increasing the need of quality annotated datas-sets as stated by Yadav and Bethard\cite{vikas} and Li et al.\cite{lijing}. These authors have provided a large overview of existing techniques for extracting and classifying named entities using machine and deep learning methods.

The problem of composing new NERC datasets has been the object of the study proposed by Lawson  et  al.\cite{lawson-etal-2010-annotating}. Manual annotation of large corpora is usually very costly, the authors propose, then, the usage of Amazon Mechanical Turk as a low-cost alternative. However, this method still depends on a specific budget for the task and can be very time-consuming. It also depends on the availability of annotators for each specific language which may be problematic if the aim is to generate large multilingual corpora. 

A generic method for extracting entities from Wikipedia articles was proposed in Bekavac and Tadić\cite{boke} and includes Multi-Word Extraction of named entities using local regular grammars. Therefore, for each targeted language, a new set of rules must be defined. Another automatic multilingual solutions has been proposed by Ni et al.\cite{DBLP:journals/corr/NiDF17} and Kim et al.\cite{kim} using either annotation projection on comparable corpora or Wikipedia metadata on parallel datasets. Both methods, however, still require manual annotations that are language-dependent and cannot be applied universally. Furthermore, Weber \& Vieira\cite{vieira} use a similar process to the one presented in this article for annotating Wikipedia texts using DBpedia information. However, their focus is on Portuguese only with a very simple NERC hierarchy.

The idea of using Wikipedia metadata to annotate multilingual corpora has also been proposed by Nothman et al.\cite{journals/ai/NothmanRRMC13} for English, German, Spanish, Dutch, and Russian. Despite the multilingual approach, it also requires manually annotated text.

\section{UNER Dataframe Description}

As mentioned in the previous section, the UNER hierarchy was introduced by Alves et al.\cite{alves2020uner}. It was built upon the 4-level NERC hierarchy proposed by Sekine\cite{sekinewebsite}, which was chosen as it presents a high conceptual hierarchy. The changes comparing both structures have been detailed by the authors. 
The proposed UNER hierarchy is also composed of 4 levels. Level 0 being the root node from which derives all the other levels. Level 1 consists of three main classes: \textit{Name}, \textit{Time Expression} and \textit{Numerical Expression}. Level 2 is composed of 29 named-entity categories which can be detailed in a third level with 95 types. Additionally, level 4 contains 129 subtypes (Alves et al.\cite{alves2020uner}).

This first version of the UNER hierarchy, therefore, encompasses 215 labels which can contain up to four levels of granularity depending on how detailed is the named-entity type. UNER labels are composed of tags from each level separated by a hyphen "-". As level 0 is the root and common for all entities, it is not present in the label. For example:

\begin{itemize}
    \item UNER label \textit{Name-Event-Natural\_Phenomenon-Earthquake} is composed of level 1 \textit{Name}, level 2 \textit{Event}, level 3 \textit{Natural Phenomenon} and level 4 \textit{Earthquake}.
\end{itemize}

The idea of using both Wikipedia data and metadata associated with DBpedia information to generate UNER annotated datasets compelled us to revise the first proposed UNER hierarchy. The main reason is that the automatic annotation process is based on a list of equivalences between UNER labels and DBpedia classes. 

By generating the list of equivalences, it was noticeable that not all UNER labels would have a DBpedia equivalent class. This is case for majority of Time and Numerical expressions. These cases will have to be dealt with by other automatic methods in future work.

Therefore, for this article, we consider version 2 of UNER, presented in the GitHub webpage of the project\footnote{https://github.com/cleopatra-itn/MIDAS/}. It is composed of 124 labels and its hierarchy is detailed in \hyperref[tab1]{Table 1}. 

\begin{table}\label{tab1}
\caption{Description of the number of nodes per level inside UNER v2 hierarchy.}
\centering
\begin{tabular}{|c|c|}
\hline
\textbf{Level} & \textbf{Number of nodes}\\
\hline
0 (root) & 1\\
1 & 3 \\
2 & 14\\
3 &  53\\
4 & 88\\
\hline
\end{tabular}
\end{table}

In the annotation process, we have decided to use the IOB format\cite{ramshaw-marcus-1995-text} as it is widely used by many NERC systems as showed by Alves et al.\cite{alves2020uner}. Therefore, each annotated entity token also receives in the beginning of the UNER label the letter “B” if the token is the first of the entity or “I” if inside. Non-entity tokens receive only the tag “O”.

\section{Data Extraction and Annotation} 
The workflow we have developed allows the extraction of texts and metadata from Wikipedia (for any language present in this database), followed by the identification of the DBpedia classes via the hyperlinks associated with certain tokens (entities) and the translation to UNER types and subtypes (these last two steps being language independent).

Once the main process of data extraction and annotation is over, the workflow proposes post-processing steps to improve the tokenization, implement the IOB format \cite{ramshaw-marcus-1995-text} and gather statistical information concerning the generated corpus.

The whole workflow is presented in detail in the project GitHub webpage together with all scripts that have been used, and that can be applied to any other Wikipedia language.

\subsection{Process description}
\begin{enumerate}
    \item \textbf{UNER/DBpedia Mapping}: This is a mapper that connects each pertinent DBpedia class with a single UNER tag. It was created by the members of the project, analysing each DBpedia class and associating it to the most pertinent UNER tag. A single extracted named entity might have more than one DBpedia class. For example, entity \textit{\textbf{2015 European Games}} have the following DBpedia classes with the respective UNER equivalences:
    \begin{itemize}
        \item \textbf{dbo:Event} -- \textit{Name-Event-Historical-Event}
        \item\textbf{dbo:SoccerTournament} -- \textit{Name-Event-Occasion-Game}
        \item \textbf{dbo:SocietalEvent} -- \textit{Name-Event-Historical-Event}
        \item \textbf{dbo:SportsEvent} --\textit{ Name-Event-Occasion-Game}
        \item \textbf{owl:Thing} -- \textit{NULL}
        
        
    \end{itemize}
    The value on the left represents a DBpedia class and its UNER equivalent is on the right side of the class. It maps all the DBpedia classes to UNER equivalent classes.
     \item \textbf{DBpedia Hierarchy}: This mapper assigns priorities to each DBpedia class. This is used to select a single DBpedia class from the collection of classes that are associated with an entity. Following are examples of classes and their priorities.
    \begin{itemize}
                \item \textbf{dbo:Event} -- 2
        \item\textbf{dbo:SoccerTournament} -- 4
        \item \textbf{dbo:SocietalEvent} -- 2
        \item \textbf{dbo:SportsEvent} -- 4
        \item \textbf{owl:Thing} -- 1
    \end{itemize}
     For entity \textbf{\textit{2015 European Games}}, the DBpedia class  \textbf{SoccerTournament}  presides over the other classes as it has a higher priority value. If the extracted entity has two assigned classes with the same hierarchy value the first from the list is chosen as the final one. All the DBpedia classes were assigned with a hierarchy value according to DBpedia Ontology\footnote{http://mappings.dbpedia.org/server/ontology/classes/}, where classes are presented in a structural order which allowed us to define the hierirchal levels.   
     
\end{enumerate}

\subsection{Main process}

The main process is schematized in the figure below and is divided into three sub-processes.

\begin{figure}[!ht]
\includegraphics[width=\textwidth]{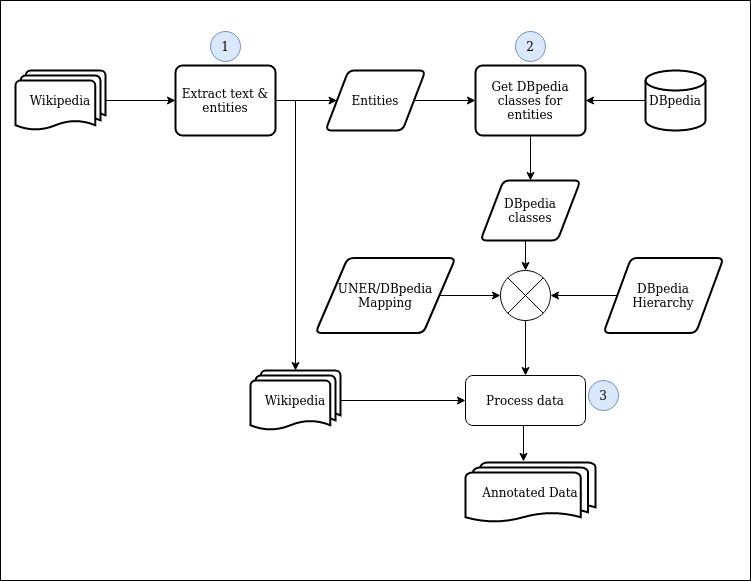}

\caption{Main process steps for Wikipedia data extraction and DBpedia/UNER annotations. Squares represent data and diamonds represent processing steps.} \label{fig1}
\end{figure}

\begin{enumerate}
    \item \textbf{Extraction from Wikipedia dumps}: For a given language, we obtain its latest dump from the Wikimedia website\footnote{https://dumps.wikimedia.org/}. Next, we perform text extraction preserving the hyperlinks in the article using WikiExtractor\footnote{https://github.com/attardi/wikiextractor}. These are hyperlinks to other Wikipedia pages as well as unique identifiers to those named-entities. We extract all the unique hyperlinks and sort them alphabetically. These hyperlinks will be referred to as named-entities henceforth.
    \item \textbf{Wikipedia-DBpedia entity linkin}g: For all the unique named-entities from the dumps, we query the DBpedia endpoint using a SPARQL query with SPARQLWrapper\footnote{https://rdflib.dev/sparqlwrapper/} to identify the various classes associated with the entity. This step produces, for each named-entity from step 1, a set of DBpedia classes it belongs to. 
    \item \textbf{Wikipedia-DBpedia-UNER back-mapping}: For every extracted named-entity obtained in step 1, we use the set of classes produced in step 2, along with a UNER/DBpedia mapping schema, to assign UNER classes to each named-entity. For an entity, all the classes obtained from the DBpedia response are mapped to a hierarchy value, a highest valued class is resolved and chosen, and then it is mapped to UNER class. For constructing the final annotation dataset, we only select those sentences that have at least one single named entity. This reduces the sparsity of annotations and thus reduces the false negatives rate in our test models. This step produces an initial tagged corpus from the whole Wikipedia dump for a specific language. 
\end{enumerate}

\subsection{Post-processing steps}
The post-processing steps correspond to three different scripts that provide:

\begin{enumerate}
  \item The improvement of the tokenization (using regular expressions) by isolating punctuation characters that were connected with words. In addition, it applies the IOB format\cite{ramshaw-marcus-1995-text} to the UNER annotations inside the text.
  \item The calculation of the following statistic information concerning the generated corpus: Total number of tokens, Number of Non-entity Tokens (tag “O”), Number of Entity Tokens (tags “B” or “I”), and Number of Entities (tag “B”). The script also provides a list of all UNER tags with the number of occurrences of each tag inside the corpus.
  \item Listing the entities inside the corpus (tokens and the corresponding UNER tag). Each identified entity appears once in this list, even if it has multiple occurrences in the corpus.
\end{enumerate}

The whole process and post-processing steps were applied to English language, generating the UNER English corpus which is described and evaluated in the following section. This baseline corpus is the base for the improvement experiments presented in later sections.  

\subsection{UNER English Corpus (Baseline)}

\subsubsection{General Information}

The English Wikipedia \cite{wiki:English_Wikipedia} is composed of 6,188,204 articles. After applying the main process of the proposed workflow, we obtained annotated text files divided into folders. The size of the English 
UNER corpus is presented in the following \hyperref[tab2]{table}.

\begin{table}\label{tab2}
\centering
\caption{Corpus Size.}
\begin{tabular}{|c|c|c|c|}
\hline
\textbf{Corpus} & \textbf{Size} & \textbf{Folders} & \textbf{Files}\\
\hline
English UNER & 3.3 GB & 172 & 17,150 \\
\hline
\end{tabular}
\end{table}

Statistical information concerning the corpus is obtained by applying the post-processing steps previously described. \hyperref[tab3]{Table 3} presents the main statistics about number of tokens and entities. Inside UNER English Corpus, 8.9\% of tokens are entities.

\begin{table}\label{tab3}
\centering
\caption{Corpora Annotation Statistics.}
\begin{tabular}{|c|c|}
\hline
 & \textbf{English UNER Corpus} \\
\hline
Total Number of Tokens & 325,395,838\\ 
Number of Non-Entity Tokens & 320,719,350 \\ 
Number of Entity Tokens & 31,676,488 \\ 
Number of Entities & 15,101,318 \\ 
Number of Different Entities & 630,519 \\ 
\hline
\end{tabular}
\end{table}

As presented in section 3, the UNER hierarchy used for annotating the English Wikipedia texts is composed of 124 different multi-leveled labels with equivalences to DBpedia classes. However, baseline UNER English corpus contains 99 different UNER tags (80\%).

As explained previously, the UNER hierarchy is composed of categories, types, and subtypes. UNER includes the most common classes used in NERC (\textit{Person}, \textit{Location}, \textit{Organization}), being more detailed (subtypes):

\begin{itemize}
  \item \textit{Person}: correspond to UNER \textit{Name-Person-Name}
  \item \textit{Location}: correspond to all subtypes inside the UNER types \textit{Name-Location}.
  \item \textit{Organization}: correspond to all subtypes inside the UNER type \textit{Name-Organization}
\end{itemize}

Therefore, it is possible to analyse the generated corpora in terms of these more generic classes.

\begin{table}\label{tab4}
\centering
\caption{Corpora Annotation Statistics in terms of number of occurrences of the most used NERC classes (and \% of all entities occurrences).}
\begin{tabular}{|c|c|}
\hline
 & \textbf{English UNER Corpus} \\ 
\hline
Person & 4,200,313 (27.8\%) \\ 
Location & 2,613,248 (17.3\%) \\ 
Organization & 3,489,813 (23.1\%) \\ 
\hline
\end{tabular}
\end{table}

These main classes correspond to 68.2\% of NEs in the generated corpus. 

\subsubsection{Qualitative evaluation}
The proposed process requires the identification of the DBpedia classes associated with the respective tokens (via hyperlink) and the translation to UNER using (UNER/DBpedia equivalences). 

An analysis of 943 entities randomly selected from UNER English Corpus has been performed to evaluate this step of the workflow. For each one, we have checked the DBpedia associated classes and the final UNER chosen tag. \hyperref[tab5]{Table 5} presents the results of this evaluation.

\begin{table}\label{tab5}
\centering
\caption{Evaluation of the annotation step: DBpedia class extraction and translation to the UNER hierarchy.}
\begin{tabular}{|c|c|c|}
\hline
\textbf{Tag Evaluation} & \textbf{Number of Occurrences} & \textbf{Percentage}\\
\hline
Correct & 797 & 85\% \\
Correct but vague & 55 & 6\%\\
Incorrect due to DBpedia & 62 & 7\%\\ 
Incorrect due to UNER association & 29 & 3\%\\
\hline
\end{tabular}
\end{table}

In the selected sample, 91\% of the entities are correctly tagged with UNER tags. Nevertheless, 6\% are associated with the correct UNER type but to a generic subtype. For example, \textbf{\textit{Bengkulu}} should be tagged as \textit{Name-Location-GPE-City} but received the tag \textit{Name-Location-GPE-GPE\_Other}. Errors may come from mistakes in the DBpedia classes associated with the tokens or due to the prioritization rules and equivalences defined between DBpedia and UNER:

\begin{itemize}
    \item \textbf{\textit{Buddhism}} is associated only to the DBpedia class \textit{EthnicGroup} and, therefore, is wrongly tagged as \textit{Name-Organization-Ethnic\_Group\_other} while it should be associated to the UNER tag \textit{Name-Product-Doctrine\_Method-Religion}.
    \item \textbf{\textit{Brit Awards}}, due to the prioritization of DBpedia class hierarchy in the choice of UNER tags, is wrongly tagged as \textit{Name-Organization-Corporation-Company} while it should receive the tag \textit{Name-Product-Award}.
\end{itemize}

\subsubsection{UNER English Golden dataset}

Beside the statistical information presented above, a sample from the generated corpus has been selected and corrected using WebAnno\cite{eckart-de-castilho-etal-2016-web} by one annotator. The sample corresponds to one entire file from the output folder and contains 519 sentences and 105 different UNER labels (out of 124 from the list of UNER-DBpedia equivalences). The annotations were done by a non-native English speaker who is a member of the project. He followed objective guidelines, and for some specific entities, research using Wikipedia were done. In cases of multi-possible assignments, a final choice was done by the annotator so that each entity would have only one label in the golden set. 

\hyperref[tab6]{Table 6} presents the evaluation results of the baseline annotations of the file used to create the Golden dataset in terms of Precision, Recall, and F1-measure, considering the mean value of all 105 labels for each metric.

\begin{table}[!ht]\label{tab6}
\caption{Precision, Recall, and F1 Measure values of UNER EN dataset considering 519 manually annotated sentences.}
\begin{center}
\begin{tabular}{|c|c|c|c|}
\hline
\textbf{Experiment} & \textbf{Precision}     & \textbf{Recall}        & \textbf{F1-measure}    \\ 
\hline
Baseline   & 61.9          & 27.2          & 37.8          \\

\hline
\end{tabular}
\end{center}
\end{table}

As explained previously, the annotation of a certain named-entity depends on the existence of hyperlinks. However, these links are not always associated with the tokens if the entity is mentioned repeatedly in the article. This may be one of the main reasons of the low value obtained for recall.

\section{Dataset Improvement}

Evaluation of the baseline annotated file using the Golden UNER English Corpus shows that the automatic annotation workflow has room for improvement, especially in terms of reducing the number of false negatives. Strategies for completing the annotation using dictionaries and knowledge graph were applied to the English Corpus. The ensemble of experiments and the evaluation is presented in the subsections below.

\subsection{Experiment Design}
Seven different experiments were conducted:

\begin{enumerate}
    \item Global Dictionary: From the whole UNER English Corpus, we have established a dictionary of entities and the respective UNER label. As the same entity may appear in the corpus with different UNER tags (due to the associated DBpedia classes), we have selected for each entity the label with the highest number of occurrences. This dictionary is then used to complete the annotations of the corpus. Only entities with length longer than 2 characters were considered and numerical entities were excluded from the dictionary. Final size of the global dictionary is of 826,371 entities.
    \item Global Dictionary only with multiple token entities: Similar to the previous experiment but in this case only entities with more than one token were considered. In total, the global dictionary is composed of 665,081 multi-token entities. 
    \item Local Dictionaries: In this setup we processed every Wikipedia dump file as a single article. Every entity in the article that is linked to UNER is cached into a local lookup dictionary with its text as the key and UNER class as the value. For every subsequent occurrence of the text in the given article we annotated the text with corresponding UNER class. We performed this step with the speculation that entities are more likely to appear within a single article than in a completely unrelated article. For example, \textit{Barrack Obama} as person is more likely to appear in an article describing him as president than as a fictional character which appears in fictional content about him.
    \item Global OEKG Dictionary: Open Event Knowledge Graph (OEKG)\footnote{http://cleopatra-project.eu/index.php/open-event-knowledge-graph/} is a multilingual event-centric resource. Its instances have specific DBpedia classes, therefore, we intersected all the entries from the global dictionary with elements from OEKG. For each entity, its associated DBpedia class from OEKG was then mapped to UNER. The global OEKG dictionary contains 128,813 entries. 
    \item Global OEKG Dictionary only with multiple tokens entities: Similar to experiment 4 only in this case only entities with more than one token were considered (110,226 entities in total). 
    \item Local Dictionaries followed by Global OEKG Dictionary: Combination of experiment 3 with completion of annotations using dictionary established for experiment 4.
    \item Local Dictionaries followed by OEKG Dictionary only with multiple tokens entities: Corpus from experiment 3 is completed using dictionary from experiment 5.
\end{enumerate}

In all experiments, dictionaries were ordered from the longest entities to the shortest ones to guarantee that preferably multi-token entities were annotated and not mono-token ones. 

\subsection{Evaluation}

The evaluation was conducted using the Golden Corpus presented previously. The baseline is the correspondent file with automatic annotations as result of the workflow described in section 4.

Golden Corpus has 105 different UNER labels, however, the baseline annotated file has only 62. For each possible label, we calculated precision, recall, and F1-measure. The IOB format\cite{ramshaw-marcus-1995-text} was applied, therefore, each UNER label can start either with "B" or "I", and non-entity tokens were tagged with "O".

From the 62 labels of the baseline, only 45 presented results different than 0. Therefore, the values present in the following \hyperref[tab7]{table} consider only these tags and represent the mean value of all the tags taken into account. \hyperref[tab7]{Table 7} presents the metrics obtained for the baseline and each one of the experiments described in the previous sub-section. 

\begin{table}\label{tab7}
\caption{Precision, Recall, and F1 Measure values of experiments for improving UNER EN dataset.}
\begin{center}
\begin{tabular}{|l|l|l|l|}
\hline
\textbf{Experiment} & \textbf{Precision} & \textbf{Recall} & \textbf{F1-measure} \\ \hline
Baseline   & 72.9      & 32.0   & 39.2       \\
1          & 32.1      & \textbf{35.7}   & 27.0       \\
2          & 47.5      & 34.8   & 34.0       \\
3          & 73.6      & 29.6   & 36.8       \\
4          & 71.1      & 33.9   & \textbf{40.8}       \\
5          & 73.0      & 33.4   & 40.5       \\
6          & 72.1      & 32.1   & 39.1       \\
7          & \textbf{74.0}      & 31.5   & 38.7       \\ \hline
\end{tabular}
\end{center}
\end{table}

Using the global dictionary (experiment 1) provides the highest value of recall (+3.7 compared to the baseline) but precision is considerably lower (-40.8). Similar situation when the global dictionary is used only with multi-token entities (experiment 2). Other experiments do not decrease precision so drastically and in some cases this metric is even increased. Recall is increased, compared to the baseline, for all experiments except for 3 and 6, 7. The usage of local dictionaries was not an effective solution for improving this evaluation metric. 

Best option, considering F1 measure, is the usage of the dictionary verified with OEKG (experiment 4). Precision is slightly lower than the baseline (-1.8) while recall and F1 measure are higher (+1.9 and +1.6 respectively). If we consider only level 3 of UNER hierarchy, the possible tags are: \textit{Disease}, \textit{Event}, \textit{Facility}, \textit{Location}, \textit{Natural Object}, \textit{Organization}, \textit{Person} and \textit{Product}

The evaluation of each experiment considering only this upper level of the UNER hierarchy is presented in \hyperref[tab8]{table 8}. The IOB format was also considered, therefore, UNER labels could be preceded by either "B" or "I" and non-entity tokens were tagged with "O".

\begin{table}\label{tab8}
\caption{Precision, Recall, and F1 Measure values of experiments for improving UNER EN dataset considering only level 3 of UNER hierarchy.}
\begin{center}
\begin{tabular}{|c|c|c|c|}
\hline
\textbf{Experiment} & \textbf{Precision}     & \textbf{Recall}        & \textbf{F1-measure }   \\ \hline
Baseline   & \textbf{76.9} & 25.1          & 34.0          \\
1          & 25.9          & \textbf{31.0} & 25.2          \\
2          & 37.2          & 27.8          & 31.1          \\
3          & 76.8          & 24.5          & 33.2          \\
4          & 74.6          & 27.3          & \textbf{36.0} \\
5          & 76.6          & 26.3          & 35.3          \\
6          & 74.6          & 26.9          & 35.5          \\
7          & 76.6          & 25.8          & 34.7          \\ \hline
\end{tabular}
\end{center}
\end{table}

In this scenario, the highest precision is from the baseline. The Best recall is obtained when the global dictionary is used (experiment 1) but, as it was observed before, in this case precision is heavily impacted compared to the baseline (-51.0). Experiment 4 is the one with the highest F1 measure, same for the previous evaluation where all UNER levels were considered. 

\section{Discussion}
In section 4 we presented the whole process for generating UNER annotated corpora and the application of this workflow to create a dataset for the English language. The evaluation of the entity’s extraction and translation to UNER (\hyperref[tab5]{table 5}) showed that 85\% of analysed entities were correctly annotated. However, errors are introduced to the dataset mainly because of the wrong association of the tokens inside Wikipedia text and DBpedia classes. Furthermore, the selection rule in our process (choosing the DBpedia class having the highest granularity inside DBpedia hierarchy) is also a source of mistakes.

The corpus generated by the proposed workflow was also evaluated using the manually annotated Golden dataset. It is noticeable that even with the reduced version of the UNER hierarchy (UNER v2, 124 labels with DBpedia equivalences) used for this task, the whole dataset was only annotated with 99 different labels, from which only 62 presented values of precision and recall different from zero in the baseline sample file when compared to the Golden. Therefore, further changes in the UNER hierarchy should be implemented, for example, Time and Numerical Expressions (already reduced in UNER v2) should be excluded in this step of the annotation and some detailed labels concerning Events, Organizations and Locations can be joined in more generic categories.

As expected, recall is much inferior to precision in all conducted evaluations. This is due to the fact that inside Wikipedia not all entities are linked to DBpedia. This problem was also encountered by Weber \& Vieira\cite{vieira} who used a similar workflow for Portuguese named-entity task with a much simpler hierarchy. The authors did not conduct any intrinsic evaluation of the generated corpus but, instead, used it to train models and evaluate results with existing Golden sets.   

Concerning the improvement experiments, the best-identified option was to use a dictionary fine-tuned from the Open Event Knowledge Graph. This graph allows to identify a more precise specific DBpedia class and therefore helps in improving recall without considerable loss in precision. In this article, we presented the use of this method as a post-processing step to complete the initial annotations of the baseline. However, the usage of OEKG information can be also implemented inside the workflow as a way of improving overall precision.

\section{Conclusions and Future Directions}
In this paper, we described an automatic workflow for generating multilingual Named-Entity recognition corpora by using Wikipedia and DBpedia data and following the UNER hierarchy. The whole process is available and can be applied to any language having Wikipedia and DBpedia. We also presented the application of the extraction and annotation method used to generate UNER English corpus. The generated dataset has been described and evaluated with a manually annotated Golden set.

Furthermore, an ensemble of experiments were conducted to improve final annotated dataset. We have identified that the best results were obtained by using a dictionary of entities with verification of the associated DBpedia class using Open Event Knowledge Graph: 76.9 for precision, 31.0 for recall and 36.0 for F1-measure. Nevertheless, there is still room for improvement in both recall and F-measure.    

In our future work, we plan to continue exploring OEKG by integrating it to extraction and annotation the workflow and not only as a post-processing step. Also, we intend to extend our corpus to other languages, especially under-resourced ones, while evaluating our workflow's performance across the languages. Moreover, to complete this intrinsic evaluation of our dataset, we plan to evaluate it extrinsically by using the generated datasets to train the machine and deep learning models.

\section{Acknowledgements}
The work presented in this paper has received funding from the European Union’s Horizon 2020 research and innovation program under the Marie Skłodowska-Curie grant agreement no. 812997 and under the name CLEOPATRA (Cross-lingual Event-centric Open Analytics Research Academy)

%
%
%
\bibliographystyle{splncs04}
\bibliography{mybibliography}
\end{document}